\documentclass[journal]{IEEEtran}
\usepackage{graphicx}
\usepackage[normalem]{ulem}
\useunder{\uline}{\ul}{}
\usepackage{subfig}
\usepackage[cmex10]{amsmath}

\usepackage{fixltx2e}
\usepackage{url}

\begin{document}
%
\title{2D-Densely Connected Convolution Neural Networks for automatic Liver and Tumor Segmentation}

\author{Krishna~Chaitanya~Kaluva,
      	 Mahendra~Khened,
         Avinash~Kori
        and~Ganapathy~Krishnamurthi
\thanks{KC. Kaluva, M. Khened, A. Kori and G. Krishnamurthi are with the Department of Engineering Design, Indian Institute of Technology, Madras,
Chennai, 600036 India email: (gankrish@iitm.ac.in).}
\thanks{Manuscript received August 13, 2017; revised August 15, 2017.}}

\markboth{Journal of TODO,~Vol.~X, No.~X, AUGUST~2017}%
{Kaluva \MakeLowercase{\textit{et al.}}: Lits_2017}
%



\maketitle

\begin{abstract}
In this paper we propose a fully automatic 2-stage cascaded approach for segmentation of liver and its tumors in CT (Computed Tomography) images using densely connected fully convolutional neural network (DenseNet). We independently train liver and tumor segmentation models and cascade them for a combined segmentation of the liver and its tumor. The first stage involves segmentation of liver and the second stage uses the first stage's segmentation results for localization of liver and henceforth tumor segmentations inside liver region. The liver model was trained on the down-sampled axial slices $(256 \times 256)$, whereas for the tumor model no down-sampling of slices was done, but instead it was trained on the CT axial slices windowed at three different Hounsfield (HU) levels. On the test set our model achieved a global dice score of 0.923 and 0.625 on liver and tumor respectively. The computed tumor burden had an rmse of 0.044. 
\end{abstract}

\IEEEpeerreviewmaketitle

\section{Introduction and Related Work}
The exponential growth in the number of medical images of different modalities used in clinical practice has led to an interest in developing automated methods for analysis of medical images. 
Computed tomography (CT scan) is a noninvasive diagnostic imaging procedure that uses a combination of X-rays and computer algorithms to generate different internal views of the body (often called slices). A CT scan shows detailed images of internal organs and structures inside the body, which also includes the bones, muscles and fat. CT scans of the abdomen and pelvis is used as a diagnostic imaging test to aid in the detection of diseases of liver, biliary tract, the small intestine and colon. CT scanning is painless, noninvasive and accurate. When compared to Magnetic Resonance Imaging (MRI), CT has wider availability, and is fast. CT images are generated using X-ray beams and the amount of X-rays absorbed by tissues at each location in the body is mapped to Hounsfield units (HU). The denser the tissue, the more the X-rays are attenuated, and the higher the number of HU. Water is always set to be 0 HU, while air is −1000 HU, and bones have values between several hundred to several thousand HU. 

Manual segmentation of the organs and tumors from medical images is a tedious task and often introduces inter-rater variability. Convolutional neural networks (CNNs) \cite{cnn} have been applied to wide variety of image classification \cite{alex,vgg,resnet} and semantic segmentation \cite{fcn, unet} tasks. In this paper, we focus on segmentation of liver and its tumors in CT images taken from thoracic to pelvis region using CNNs. Our network's architecture for segmentation task is inspired from DenseNet \cite{densenet,tiramisu}. DenseNet connects each layer to every other layer in a feed-forward fashion by concatenation of all feature outputs. The output of the $l^{th}$ layer is defined as  
\begin{equation}
\label{eq:dense}
x_{l}=H_{l}([x_{l-1},x_{l-2}, \cdots, x_0])
\end{equation}
where $x_{l}$ represents the feature maps at the $l^{th}$ layer and $[\cdots]$ represents the concatenation operation. In our case, $H_{l}$ is the layer comprising of Batch Normalization (BN) \cite{bn}, followed by Exponential Linear Unit (ELU) \cite{elu}, a convolution and dropout \cite{dropout}. This kind of connectivity pattern aids in reuse of features and allows implicit deep supervision during training and thus substantially reduces the number of parameters while maintaining good performance, which is ideal in scenarios with limited data. The output dimension of each layer has $k$ (growth rate parameter) feature maps. The number of feature maps in DenseNet grow linearly with depth. A Transition Down (TD) layer in DenseNet is introduced for reducing spatial dimension of feature maps which is accomplished by using a $1\times1$ convolution (depth preserving) followed by a $2\times 2$ max-pooling operation. A denseblock refers to concatenation of new feature maps created at a given resolution.

\section{Our Method}
CT Windowing is a technique frequently used in the evaluation of CT scans for the purpose of enhancing contrast of particular type of tissue or the abnormality type being examined. The abdominal CT images of a patient comprises of various organs such as liver, spleen, gal bladder, etc. Anatomically, the HU range of liver is $60\pm6$. Our method is based on 2-stage cascaded approach \cite{patrick} for segmentation of liver and its tumors from HU windowed CT volumes. In this method we first train the liver model for the task of liver segmentation. Since the shape (contour) and texture of liver is simple when compared to its tumors, the CT images after windowing were down-sampled to half their original size ($512 \times 512$) and then used for training. This helped in reducing the computation required for the training liver model and also helps in faster segmentation of liver.
The tumor model was 3-channel input and was trained independently on full-sized CT images of liver windowed at $3$ different levels having different window widths. During prediction, the liver segmentation from first stage in the cascade (liver model) aids the second stage (tumor model) by precisely localizing on liver regions in CT images to produce combined predictions of liver and tumor segmentations. 
\subsection{DATASET AND PREPROCESSING}
The models were trained and tested on the LITS MICCAI-2017 challenge dataset which comprised of $200$ contrast enhanced CT images taken at different phases (mostly venous phase) with only a few cases with anomalies like fatty liver, cirrhosis liver, calcification in the liver, etc. Out of $200$ CT scans, for the $130$ scans radiologist hand-drawn ground-truths were given for training the model and the rest $70$ were used for testing by the challenge organizers. We divided the $130$ training dataset into train:– $90$ volumes, validation:– $26$ volumes, test:– $14$ volumes.

For liver model, the following pre-processing techniques were done on the CT volumes in the order specified:
\begin{enumerate}
\item HU values are windowed to the range of [-100,300].
\item $0-1$ min-max normalization on the entire volume.
\item Down-sample the slices from $512 \times 512$ to $256 \times 256$. 
\end{enumerate}

For tumor model, the following pre-processing techniques were done on the CT volumes in the order specified:
\begin{enumerate}
\item 3 different HU windowing ranges ([0,100], [-100,200], [-100,400]) were used to produce 3 images.
\item $0-1$ min-max normalization on the entire volume.
\end{enumerate}
In most of the CT volumes (in the challenge dataset), the liver and tumor slices comprised of a small fraction of the total volume. So in order to address the data imbalance, the liver model was trained only on liver slices and additional 10 slices were taken above and below the liver. Similarly, the tumor model was trained only on tumor slices and with additional 5 slices above and below the tumor. 

\subsection{Liver Model}
\subsubsection{Network Architecture}\hspace*{\fill} \\
\begin{figure}
\begin{center}
 \includegraphics[width=0.35\textheight]{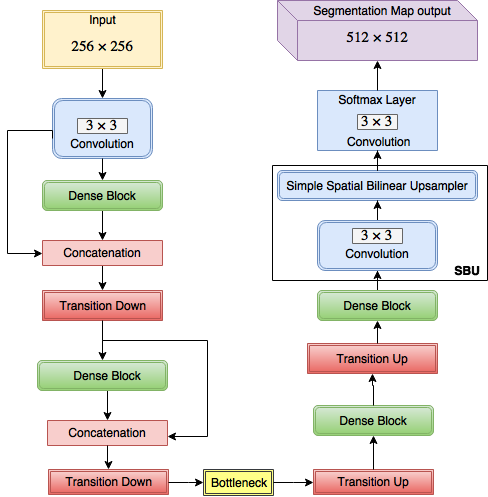}
\end{center}
  \caption{Network Architecture of liver model.}
  \label{fig:arch_liver}
\end{figure}
Fig.(\ref{fig:arch_liver}) illustrates the schematic diagram of the proposed  network for liver segmentation. Our proposed network doesn’t have skip connections. The up-sampling and down-sampling paths uses fully convolutional denseblocks. Each denseblock comprises of 4 layer blocks and each layer in the denseblock is sequentially composed of BN $\rightarrow$ ELU $\rightarrow$ $3\times3$ convolution layer. The first Dense-Block was prefixed with a layer comprising of several convolution filters of size $3\times3$ on the input images. In the down-sampling path, the input to a dense block was concatenated with its output, leading to a linear growth of the number of feature maps. The Transition-Down block (TD) consists of BN $\rightarrow$ ELU $\rightarrow$ $1\times1$ convolution and $\rightarrow$ $2\times2$ max-pooling layers. The last layer of the down-sampling path is referred to as Bottleneck.\\
In the up-sampling path, the input of a Dense-Block is not concatenated with its output. Transition-Up (TU) block consists of spatial-bilinear up-sampler followed by BN $\rightarrow$ ELU $\rightarrow$ and $\rightarrow$ $3 \times3$ convolution layer. During our training we found that by using spatial bilinear up-sampler followed by convolution yielded better results compared to transposed convolution (deconvolution) operation with a stride of $2$. Our thesis is that the network also learns better way of up-sampling the outputs which resulted in better predictions. To get the final segmentation map of size of $512 \times 512$ a simple spatial bilinear up-sampling block (SBU) is added in the penultimate layer. The feature maps of the hindmost up-sampling component was convolved with a $3\times3$ convolution layer followed by a soft-max layer to generate the final segmentation. To prevent over-fitting, a dropout rate of $0.2$ was implemented following each convolution layer.\\
\begin{table}
\parbox{.2\linewidth}{
\centering
\caption*{\textbf{Layer}}
\begin{tabular}{|c|ll}
\cline{1-1}
Batch Normalization      &  &  \\ \cline{1-1}
Exponential Linear Unit         &  &  \\ \cline{1-1}
$3 \times 3$ Convolution &  &  \\ \cline{1-1}
Dropout $p = 0.2$        &  &  \\ \cline{1-1}
\end{tabular}
}
\hfill
\parbox{.2\linewidth}{
\centering
\caption*{\textbf{TD}}
\begin{tabular}{|c|ll}
\cline{1-1}
Batch Normalization      &  &  \\ \cline{1-1}
Exponential Linear Unit         &  &  \\ \cline{1-1}
$1 \times 1$ Convolution &  &  \\ \cline{1-1}
Dropout $p = 0.2$        &  &  \\ \cline{1-1}
$2 \times 2$ Max Pooling        &  &  \\ \cline{1-1}
\end{tabular}
}
\hfill
\parbox{.2\linewidth}{
\centering
\caption*{\textbf{TU}}
\begin{tabular}{|c|ll}
\cline{1-1}
Spatial Bilinear\\ Up-sampler \\
$3 \times 3$ Convolution\\
Batch Normalization                
&  &  \\ \cline{1-1}
\end{tabular}
}

\caption{Building blocks of the network architecture. From left to right: layer used in the model, Transition Down (TD) and Transition Up (TU).}
\label{tab:bloc}
\end{table}
Table (\ref{tab:bloc}) summaries the individual blocks of network architecture.

It was observed that by using Exponential Linear Unit (ELUs) instead of Rectified Linear units (ReLUs) led to faster convergence.\\

\subsubsection{Loss function}
\begin{figure}[h]
\subfloat[Ground Truth Segmentation]{\includegraphics[width=1.8in]{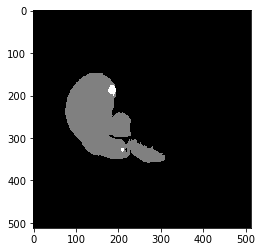}}
\subfloat[Weight Map]{\includegraphics[width=1.8in]{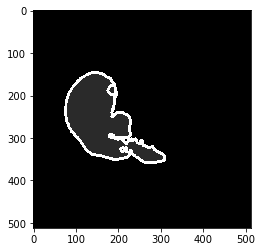}}
\caption{The figure shows the weight-map generation from the ground-truth label image}
\label{fig:wmap}
\end{figure}

\hspace*{\fill} \\
Liver being the largest organ in the body, the class imbalance shown in the CT volume is minimal, but contouring of the liver borders from its neighboring organs in CT images is generally not precise. Hence in-order to predict the edges precisely, weight-maps were generated from the given ground truth segmentation (see Figure \ref{fig:wmap}), edges of the liver were weighed higher relative to the interior regions of the liver. These weight-maps were used during training for minimizing the spatially-weighted-cross-entropy loss function. The use of weight-mapping leads to adding a heavy penalty in cost function for predicting imprecise liver contours during training the model.

\begin{equation}
\label{eq:loss_liver}
\begin{split}
total\_loss &= spatially\_weighted\_cross\_entropy\_loss + \\
 			& L2\_loss
\end{split}
\end{equation}
In our experiments we define an epoch to be completed when a specified number of iterations are executed on the randomly sampled batches of data from the train and validation set. In this experiment the number of iterations were $1000$ and $250$ for train and validation datasets respectively. 

The parameters of the network were optimized so as to minimize the $total\_loss$, Equation. (\ref{eq:loss_liver}). The liver model was trained with a batch size of $4$ for $80$ epochs with a learning rate of $10^{-4}$ and L2 weight-decay of $10^{-6}$ using ADAM optimizer\cite{adam}.\\

\subsubsection{Post processing}\hspace*{\fill} \\
    The outputs of the Liver network were subjected to the following post-processing methods sequentially:
    \begin{enumerate}
    \item Morphological binary erosion
    \item Get the largest connected component
    \item Morphological binary dilation
    \end{enumerate}
    By applying these techniques, we were able to remove some false positives like spleen or heart that might be very closely attached to the liver.

\subsection{Tumor Model}
\subsubsection{Network Architecture}\hspace*{\fill} \\

The network architecture for tumor model was similar to liver model, except that it accepted 3-channel inputs and there was no simple spatial bilinear up-sampling block (SBU) as in the liver model. The 3 channels were feed with the same slice of the CT volume, but each channel’s HU - windowing was at 3 different levels having different window widths. It was observed that by providing 3 different HU windowed channels, the model is able to learn the tumor and its boundary very well compared to a single channel input. To prevent over-fitting a dropout rate of $0.2$ was implemented.\\

\subsubsection{Loss function}\hspace*{\fill} \\
The shape of tumors is diffusive, inhomogeneous and sparsely located in the CT volumes. This leads to class imbalance in the dataset, thereby making it hard for the network to learn the intricate features of the tumor. Hence in order to reduce the class imbalance we employ two strategies:
\begin{itemize}
\item Use of weight-maps with tumor portion being weighed very high compared to background. The weight-maps are used for calculation of spatially-weighted-cross-entropy loss.
\item Use weighted combination of two loss functions: spatially-weighted-cross-entropy loss and a loss function based on dice overlap co-efficient.
\end{itemize}

\par The dice co-efficient is an overlap metric used for assessing the quality of segmentation maps. The dice coefficient between two binary volumes can be written as:
\begin{equation}
\label{eq:dice}
DICE =\frac{2\sum_{i}^{N}p_ig_i}{\sum_i^Np_i^2+\sum_i^Ng_i^2}
\end{equation}
where the sums run over the $N$ voxels, of the predicted binary segmentation volume $p_i \in P$ and the ground truth binary volume $g_i \in G$

The parameters of the network were optimized so as to minimize the $total\_loss$, Equation. (\ref{eq:loss_tumor}).
\begin{equation}
\label{eq:loss_tumor}
\begin{split}
total\_loss &= \lambda (spatially\_weighted\_cross\_entropy\_loss) + \\ 
& \gamma (1-dice\_loss) + L2\_loss
\end{split}
\end{equation}
where $\lambda$ and $\gamma$ are empirically assigned weights to individual losses. During training it was observed that the dice loss allowed higher overlap scores than when trained with the loss function based on the cross entropy loss alone. In this work we set $\gamma = 0.5$ and $\lambda = 0.5$.\\

\subsubsection{Post processing}\hspace*{\fill} \\
    The predictions of the tumor network were masked with the background taken from the liver prediction, hence the final tumor predictions were only inside the liver.

\section{Conclusion}
With the pre-processing, network architecture and post processing steps discussed in this paper we were able to achieve an average liver dice of 0.93 on the 14 test volumes (mentioned in the dataset split up in the beginning) and an average tumor dice of 0.45 on the 14 test volumes. 
Our liver predictions have a smoother surface (3D volume) and precise edges compared to the liver predictions from a U-net whose predictions have ridges on the surface that arise because of very finer slice wise predictions that result because of the skip connections. The true positives of tumor prediction have almost more than 50\% overlap with the ground truth but the model predicts a lot of false positives like gallbladder edges, diaphragm, vessels etc. 
The results of our proposed 2-stage cascaded model's predictions on the 70 CT test volumes is summarized in Tabels \ref{tab:res1} \& \ref{tab:res2}.

\begin{table}[!h]
\centering
\begin{tabular}{|c|c|c|}
\hline
{\ul \textbf{Metrics}}   & \textbf{\begin{tabular}[c]{@{}c@{}}Computed LIVER \\ SEGMENTATION metrics\end{tabular}} & \textbf{\begin{tabular}[c]{@{}c@{}}Computed LESION \\ SEGMENTATION metrics\end{tabular}} \\ \hline
\textbf{voe}             & 0.150                                                                                   & 0.411                                                                                    \\ \hline
\textbf{dice global}     & 0.923                                                                                   & 0.625                                                                                    \\ \hline
\textbf{dice}            & 0.912                                                                                   & 0.725                                                                                    \\ \hline
\textbf{rmsd}            & 9.682                                                                                   & 2.070                                                                                    \\ \hline
\textbf{rvd}             & -0.008                                                                                  & 19.705                                                                                   \\ \hline
\textbf{assd}            & 6.465                                                                                   & 1.441                                                                                    \\ \hline
\textbf{jaccard}         & 0.850                                                                                   & 0.589                                                                                    \\ \hline
\textbf{dice\_per\_case} & 0.912                                                                                   & 0.492                                                                                    \\ \hline
\textbf{msd}             & 45.928                                                                                  & 7.515                                                                                    \\ \hline
\end{tabular}
\caption{Results of segmentation metrics on the test set comprising of 70 CT volumes}
\label{tab:res1}
\end{table}

\begin{table}[!h]
\centering
\begin{tabular}{|c|c|}
\hline
\textbf{Computed LESION DETECTION metrics} & \textbf{Computed TUMOR BURDEN} \\ \hline
recall: 0.348                              & rmse: 0.044                    \\ \hline
precision\_greater\_zero: 0.211            & max: 0.194                     \\ \hline
precision: 0.117                           &                                \\ \hline
recall\_greater\_zero: 0.628               &                                \\ \hline
\end{tabular}
\caption{Results of lesion detection and tumor burden estimation metrics on the test set comprising of 70 CT volumes}
\label{tab:res2}
\end{table}

\appendices
\section{Segmentation Results}
\begin{figure}[h]
\begin{center}
 \includegraphics[width=0.35\textheight]{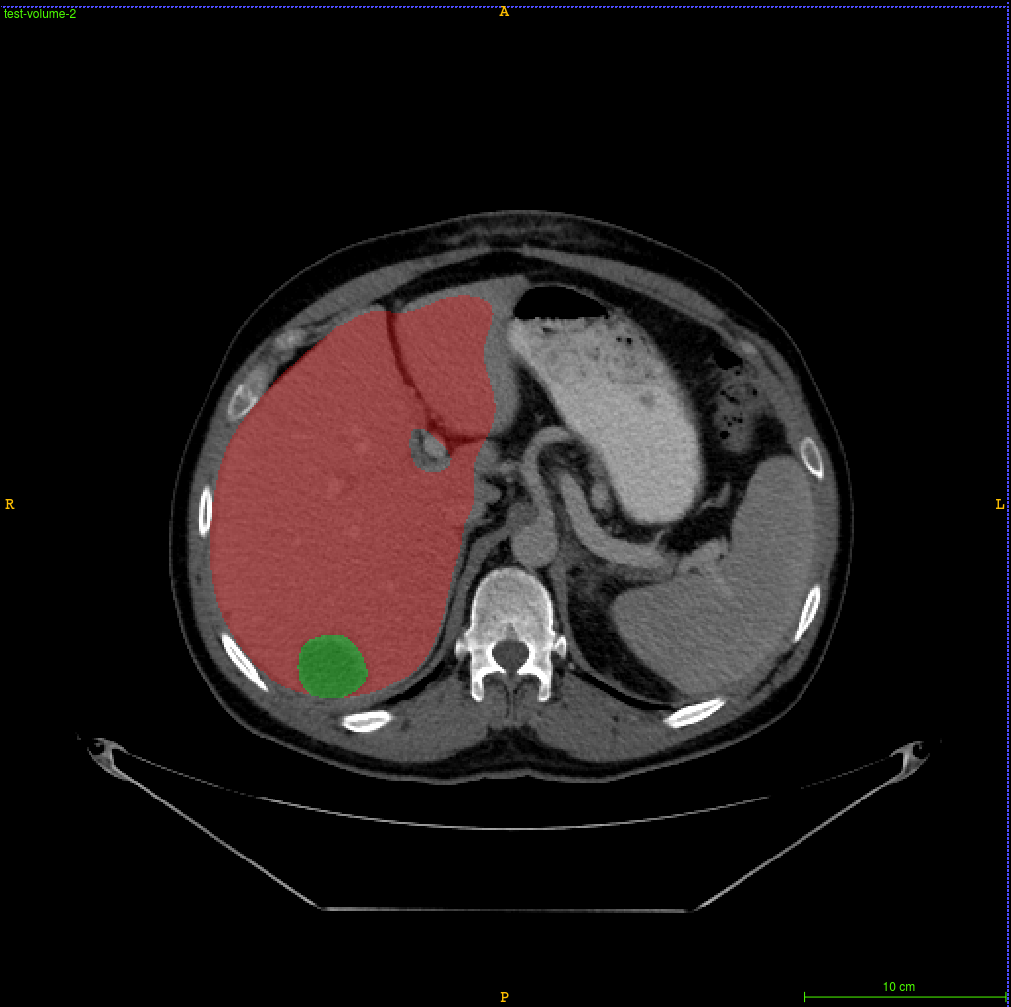}
\end{center}
 \caption{The figure shows the segmentation results of liver and tumor in a CT slice of test dataset. The color red indicates liver and green indicates tumor. The proposed model sometimes mis classifies darker vessels as tumors.}
\end{figure}

\end{document}